\documentclass[lettersize,journal]{IEEEtran}
\usepackage{amsmath,amsfonts}
\usepackage{array}
\usepackage[caption=false,font=normalsize,labelfont=sf,textfont=sf]{subfig}
\usepackage{textcomp}
\usepackage{stfloats}
\usepackage{url}
\usepackage{verbatim}
\usepackage{graphicx}
\usepackage{cite}
\usepackage{pifont}%
\usepackage{amssymb}%
\usepackage{ulem}
\usepackage{algorithmic}
\usepackage{algorithm}

\usepackage[table]{xcolor}

\usepackage{hyperref}
\usepackage{url}
\usepackage[utf8]{inputenc} 
\usepackage[T1]{fontenc}    
\usepackage{xr-hyper} 
\usepackage{enumitem}
\usepackage{cuted}

\usepackage{booktabs}       
\usepackage{amsfonts}       
\usepackage{nicefrac}       
\usepackage{microtype}      

\usepackage{graphicx}
\usepackage{caption}
\usepackage{subcaption}
\usepackage{float}
\usepackage{wrapfig}
\usepackage{multirow}

\begin{document}

\title{FeatureSORT: Essential Features for Effective Tracking}
    
\author{Hamidreza Hashempoor, Rosemary Koikara, Yu Dong Hwang
\thanks{H. Hashempoor, R. Koikara and Y. Hwang are with Pintel Co. Ltd., Seoul, South Korea (e-mail: \{Hamidreza, ~Rosemary, ~Ydhwang\}@pintel.co.kr).}
}



\maketitle

\begin{abstract}
We introduce FeatureSORT, a simple yet effective online multiple object tracker that reinforces the DeepSORT baseline with a redesigned detector and additional feature cues. In contrast to conventional detectors that only provide bounding boxes, our modified YOLOX architecture is extended to output multiple appearance attributes, including clothing color, clothing style, and motion direction, alongside the bounding boxes. These feature cues, together with a ReID network, form complementary embeddings that substantially improve association accuracy. Furthermore, we incorporate stronger post-processing strategies, such as global linking and Gaussian Smoothing Process interpolation, to handle missing associations and detections. During online tracking, we define a measurement-to-track distance function that jointly considers IoU, direction, color, style, and ReID similarity. This design enables FeatureSORT to maintain consistent identities through longer occlusions while reducing identity switches. Extensive experiments on standard MOT benchmarks demonstrate that FeatureSORT achieves state-of-the-art online performance, with MOTA scores of 79.7 on MOT16, 80.6 on MOT17, 77.9 on MOT20, and 92.2 on DanceTrack, underscoring the effectiveness of feature-enriched detection and modular post-processing in advancing multi-object tracking.
\end{abstract}


\section{Introduction}

\IEEEPARstart{G}{oal} of multi-object tracking (MOT) is to detect and consistently track all objects across video frames, a task that plays a central role in video understanding. However, the complexity of real-world videos containing many objects often leads to inaccurate bounding boxes, as detectors remain vulnerable to imperfect predictions. Moreover, MOT methods \cite{zhang2021fairmot} must balance true and false positives by filtering out low-confidence detections and retaining only reliable ones \cite{luiten2021hota}. Thanks to the rapid progress in detection modules in recent years, this issue has become more manageable, establishing tracking-by-detection (TBD) as the most effective MOT paradigm. Since MOT depends heavily on both detection and re-identification (ReID), most TBD approaches uses state-of-the-art detectors and ReID models to improve performance \cite{luo2019strong, hashempour2020data, ghalamzan2021deep, hashempoor2025glance}. Typically, these methods first extract appearance and motion features of objects, then apply a matching algorithm to associate detections with existing tracklets.

\begin{figure}[t]
  \centering
  \includegraphics[width=\linewidth]{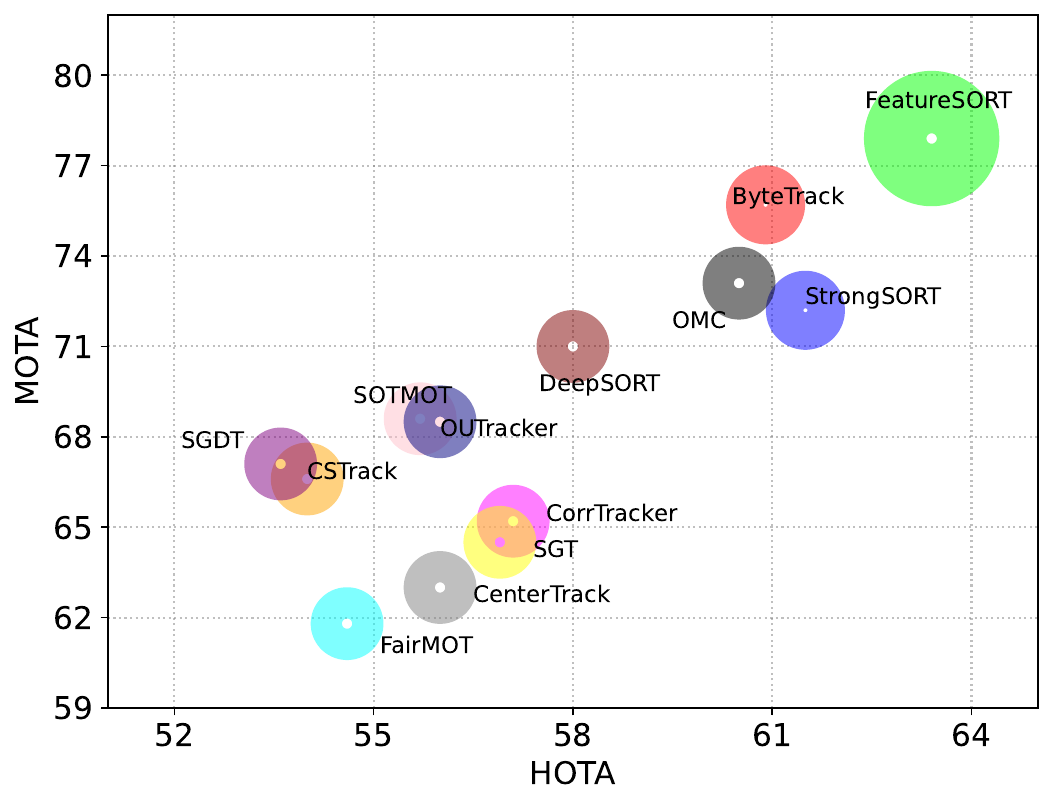} \\
  \vspace{-0.1cm} 
  \includegraphics[width=\linewidth]{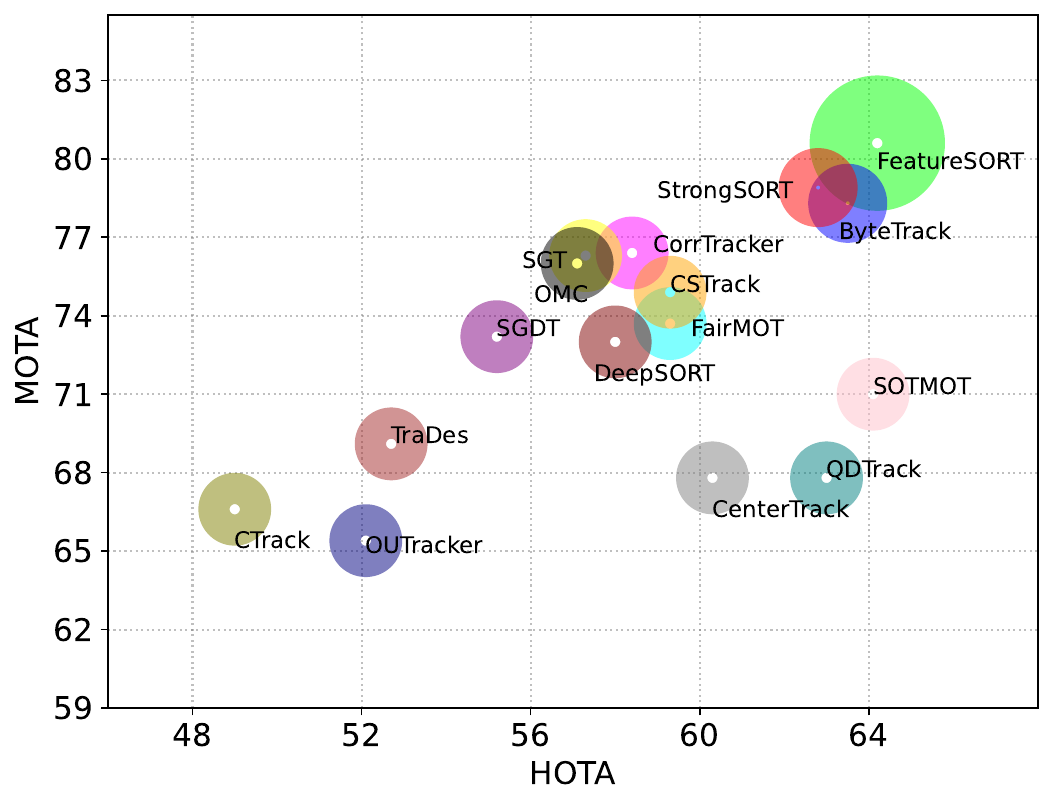}
  \caption{MOTA–HOTA comparisons of SOTA trackers with FeatureSORT (ours) on the MOT20 (top) and MOT17 (bottom) test sets, without offline post-processing.}
  \label{fig:MOSTSOTA}
\end{figure}

Building on this foundation, we present FeatureSORT, a simple yet effective MOT baseline. Our approach extends the classic DeepSORT tracker \cite{wojke2017simple}, one of the earliest methods to integrate deep learning into the MOT task. Despite criticism of DeepSORT for its relatively outdated tracking algorithm, it remains a simple, explainable, and widely adopted baseline with strong potential to be enhanced through modern features and inference strategies.
Concretely, we first integrate a redesigned YOLOX detector \cite{luo2019strong}, which not only provides stronger detection performance but also predicts additional appearance attributes of each target. In particular, the detector is trained to output clothing color and style using public pedestrian attribute datasets such as UPAR \cite{specker2023upar} and MSP60K \cite{jin2025pedestrian}, and to estimate movement direction using the MEBOW dataset \cite{wu2020mebow}. Second, beyond the standard ReID embeddings, we equip DeepSORT with separate feature modules, each representing a specific characteristic of pedestrian appearance. While ReID networks are effective for extracting generic embeddings, these representations may fail to capture finer attributes such as clothing color, style, or motion direction. Third, we adopt advanced post-processing methods from recent works to further improve DeepSORT’s overall tracking accuracy.
Enhancing DeepSORT with these components forms the proposed FeatureSORT, which achieves SOTA performance on standard benchmarks such as MOT16, MOT17, and MOT20 and DanceTrack .
Overall, the motivations for FeatureSORT can be summarized as follows:
\begin{enumerate}
\item To overcome the limitations of existing TBD trackers by introducing a feature-enriched detection framework, where the detector outputs not only bounding boxes but also appearance attributes such as clothing color, clothing style, and motion direction.
\item To establish a modular and versatile baseline that integrates these feature cues with ReID embeddings, making it easily extensible and adaptable to different tracking methodologies and environments.
\item To demonstrate that the strategic integration of advanced inference and post-processing methods can substantially enhance tracking accuracy, reduce identity switches, and improve robustness under occlusion.
\end{enumerate}

In the context of MOT, two predominant challenges are \textit{missing association} and \textit{missing detection}, as identified by Du et al. \cite{du2023strongsort}. \textit{Missing association} occurs when the same object is mistakenly split across multiple trajectories, a problem particularly common in online trackers that lack global context. \textit{Missing detections}, also known as false negatives, arise when objects are misclassified as background due to factors such as occlusion or low resolution. To address these two challenges, we incorporate dedicated post-processing strategies into FeatureSORT.

To mitigate the problem of missing association, we adopt a global linking strategy similar to GIAOTracker \cite{du2021giaotracker}. After the online tracking stage, trajectories may be fragmented into short tracklets due to occlusions or detection errors. Our global link module employs an improved ResNet50-TP model to encode tracklet appearance features and associates them using spatial and temporal costs. Although computationally intensive, this process is offline and thus does not affect real-time operation.

In contrast, to handle missing detections, we employ interpolation techniques. While linear interpolation has been widely used \cite{hashempoor2025deep}, it ignores motion information and can yield inaccurate estimates. To improve accuracy, we integrate a Gaussian Smoothing Process (GSP) \cite{schulz2018tutorial, hashempoorikderi2024gated}, which adjusts each observation by considering both past and future frames, leading to smoother and more reliable trajectories.
Together, global linking and GSP interpolation act as lightweight, plug-and-play post-processing modules that are model-independent. In our experiments, we show that applying these methods to FeatureSORT further enhances tracking robustness.

While these post-processing strategies further boost robustness, they are typically applied offline. Since many real-world applications demand real-time tracking, we place greater emphasis on the online setting, where post-processing may not be feasible. Accordingly, we present most of our results in the online scenario. Extensive experiments demonstrate that the proposed feature modules themselves already yield significant improvements in FeatureSORT. Figure \ref{fig:MOSTSOTA} illustrates MOTA–HOTA comparisons against SOTA online trackers (without offline post-processing) on the MOT17 and MOT20 test sets. The contributions of our work are as follows:

\begin{itemize}
    \item We propose FeatureSORT, an extension of DeepSORT that integrates a redesigned YOLOX detector. Unlike conventional detectors, it outputs not only bounding boxes but also multiple appearance attributes (clothing color, clothing style, motion direction), providing richer cues for robust association. This design establishes FeatureSORT as both a high-performing tracker and a versatile baseline for future MOT research.

    \item To address common challenges of missing associations and missing detections, we introduce Global Linking and GSP module. These lightweight, offline post-processing strategies can be seamlessly plugged into FeatureSORT or other trackers to improve trajectory continuity.  

    \item We conduct extensive experiments on MOT16, MOT17, MOT20, and DanceTrack, demonstrating that FeatureSORT achieves SOTA online performance (MOTA: 79.7, 80.6, 77.9, and 92.2 respectively), while also reducing identity switches and improving occlusion handling.  
\end{itemize}

The remainder of this paper is organized as follows: Section~\ref{sec:rel_Works} reviews related work. Section~\ref{sec:deepsort} outlines the background. Section~\ref{sec:featuresort} presents FeatureSORT and its modules. Section~\ref{sec:results} details the experimental setup and results. Section~\ref{sec:discussion} discusses limitations, and Section~\ref{sec:conclusion} concludes the paper.

\section{Related Work}
\label{sec:rel_Works}

In the domain of multi-object tracking (MOT), tracking-by-detection (TBD) has become the dominant paradigm. These models first detect objects in each frame and then associate the detections across time to form trajectories. The success of a TBD tracker depends largely on the association mechanism, which must handle challenges such as occlusion, fast motion, and appearance or illumination changes. To better illustrate the development of this field, we group TBD methods into three categories based on the type of information for association.

\textbf{Location and Motion Information Based Methods.} \hspace{0.2cm}
SORT \cite{bewley2016simple} uses a Kalman Filter to predict the next object location and then assigns detections to tracklets based on overlap and the Hungarian matching algorithm. IoU-Tracker \cite{bochinski2017high} simplifies this process by relying solely on the intersection-over-union (IoU) between the last tracked location and current detections, without motion prediction. Both methods are widely used in practice due to their simplicity and efficiency, but they perform poorly in crowded scenes or with fast-moving objects.
To improve robustness, more sophisticated approaches have been explored. For example, Chu et al. \cite{chu2019online} incorporate single-object tracking modules to refine object locations, though these methods are too slow for multi-object scenarios. Zhang et al. \cite{zhang2020long} propose a motion evaluation network that learns long-term tracklet dynamics to improve association. More recently, MAT \cite{han2022mat} extends SORT by modeling camera motion and employing dynamic windows for long-range association.

\textbf{Appearance Information Based Methods.} \hspace{0.2cm}
Many recent approaches \cite{ zhou2018online, wojke2017simple} rely on cropping detection bounding boxes and feeding them into Re-ID networks \cite{luo2019strong} to extract embeddings. These embeddings are then compared using a similarity metric, and detections are matched to tracklets via assignment algorithms. Such appearance-based association is particularly effective in handling fast motion and occlusions, as object features remain relatively stable over time, allowing lost tracks to be re-initialized.
To improve the robustness of appearance features, several works have proposed enhancements. For instance, Sadeghian et al. \cite{sadeghian2017tracking} introduce an online appearance learning method to adapt to changes over time, while Tang et al. \cite{tang2017multiple} incorporate body-pose cues into the embeddings. MOTDT \cite{chen2018real} further combines IoU with appearance cues in a hierarchical association strategy, using location information when appearance features are unreliable. More sophisticated assignment mechanisms are also explored, such as RNNs to model association dependencies over time \cite{fang2018recurrent}.

\textbf{Offline Methods.} \hspace{0.2cm}
Offline, or batch, methods perform global optimization over entire sequences once all detections are available, often achieving superior accuracy compared to online approaches \cite{zhang2008global, berclaz2011multiple, braso2020learning}. Zhang et al. \cite{zhang2008global} formulate MOT as a min-cost flow problem on a graph where nodes represent detections across frames, enabling efficient global association. Berclaz et al. \cite{berclaz2011multiple} also treat data assignment as a flow optimization task, using a K-shortest paths algorithm to speed up computation.
More recent approaches employ learning-based global optimization. MPNTrack \cite{braso2020learning} introduces graph neural networks to perform end-to-end association across entire sequences. Yo et al. \cite{yu2022relationtrack} cast MOT as a lifted disjoint paths problem, allowing for long-range temporal interactions and substantially reducing identity switches. These offline methods demonstrate the advantage of incorporating full-sequence context, though their reliance on future frames limits applicability to real-time scenarios.

\begin{figure*}[t]
\vspace{-0.7cm}
\centering
\includegraphics[height=7cm, width=15.2cm]{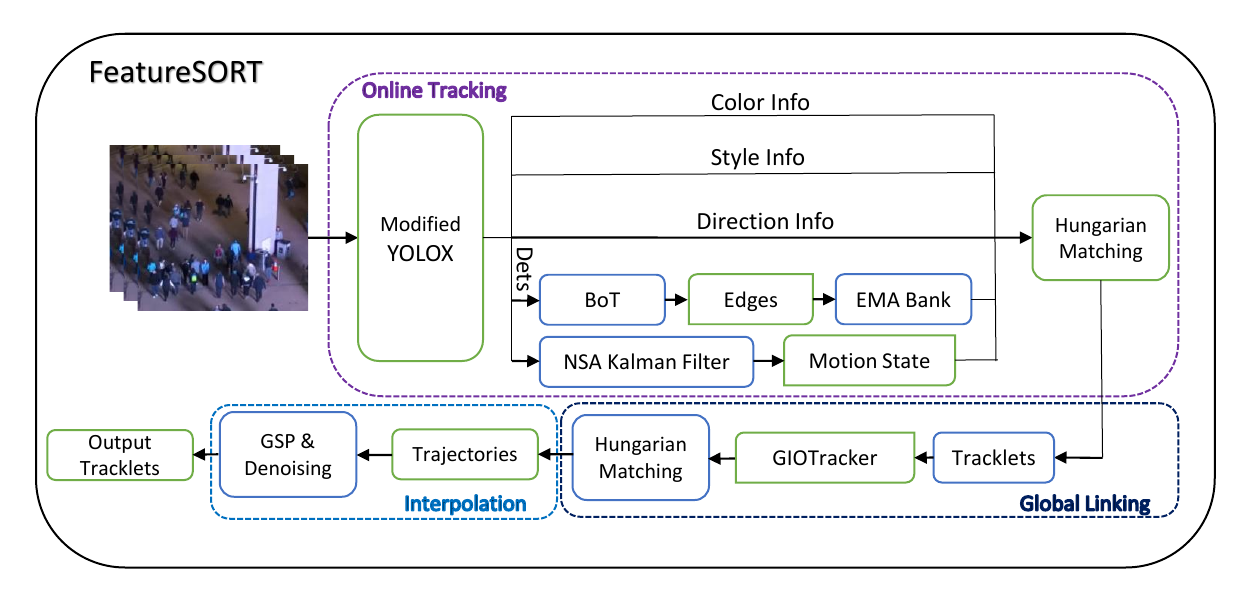}
\caption{ Schematic diagram of FeatureSORT.}
\setlength{\intextsep}{-10pt}
\vskip -0.2in
\label{fig:feature_sort}
\end{figure*}

\section{Background }
\label{sec:deepsort}
In this section, we provide an overview of the DeepSORT approach, which leverages two distinct branches to extract information: one dedicated to appearance and the other to motion. For each frame, the appearance branch uses an embedding network, pre-trained on the MARS person re-identification dataset, to extract a feature vector for each detection. DeepSORT maintains a feature bank $\mathcal{E}^{(i)}$ for each tracklet $i$, storing the most recent embeddings. For a new detection $j$ with feature vector $\mathbf{f}^j$, the appearance similarity is computed as the maximum cosine similarity to the tracklet’s feature bank:

\begin{equation}
\label{eq:cosine_dist}
\mathcal{L}_\text{ReID}(i,j) = \max_{\mathbf{e}^{(i)} \in \mathcal{E}^{(i)}}   \langle \mathbf{e}^{(i)}, \mathbf{f}^j \rangle .
\end{equation}
This distance is later incorporated into the overall matching cost for data association.

The motion branch in DeepSORT relies on a Kalman filter to estimate the positions of tracklets across frames. The Kalman filter alternates between state prediction (estimating the expected location and velocity of a target) and state update (refining the estimate using new detections). Based on these predictions, DeepSORT computes the Euclidean distance between predicted track positions and new detection bounding boxes, which serves as a motion cost for data association. To perform assignment, DeepSORT adopts a matching cascade strategy, which prioritizes recently observed tracklets, and solves each subproblem using the cascade algorithm.

\section{ Approach}
\label{sec:featuresort}

In this section, we present FeatureSORT, a multi-object tracking framework designed to combine strong detection with rich feature cues for robust online tracking. The detector is redesigned to not only provide accurate bounding boxes but also predict additional appearance attributes, including clothing color, clothing style, and movement direction. These outputs are integrated into dedicated feature modules, and a feature-aware distance function is introduced to jointly exploit ReID embeddings and attribute-based similarities for reliable association. In addition, inference and post-processing strategies are incorporated to improve robustness under occlusion and missing detections. The overall architecture of FeatureSORT is illustrated in Figure \ref{fig:feature_sort}.


\subsection{Feature-Enriched Detector}
\begin{figure}[t]
    \vspace{-0.7cm}
    \centering
    \includegraphics[width=\linewidth, keepaspectratio]{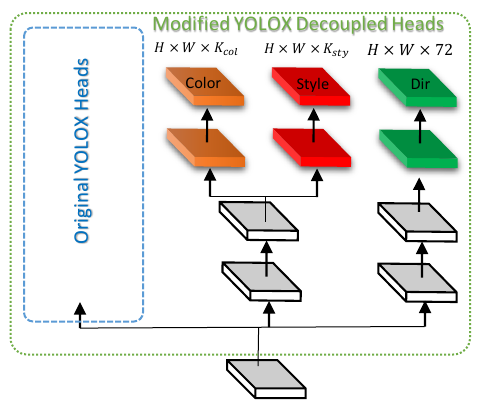}
    \caption{ Modified structure of the decoupled head of the YOLOX.}
    \setlength{\intextsep}{-10pt}
    \vskip -0.1in
    \label{fig:yolox}
\end{figure}

We adopt YOLOX \cite{ge2021yolox} as the backbone detector and redesign it to serve not only as a stronger localization model but also as a predictor of multiple target attributes. In addition to bounding boxes, our detector is trained to output appearance cues such as clothing color, clothing style, and movement direction, which provide complementary information for robust association.
To achieve this, we extend the decoupled head of YOLOX, which originally contains separate branches for localization and classification. Two additional heads are added: one for learning embeddings used in color and style classification, and another for direction classification. To preserve the detection accuracy of the backbone, localization head, and classification head, these components are frozen after the initial detection training stage, which is trained on a combination of MOT15, MOT16, MOT17, MOT20, DanceTrack, COCO, and ETHZ \cite{ess2008mobile} datasets. The newly added feature heads are then trained separately with supervision from public pedestrian attribute datasets, including UPAR \cite{specker2023upar}, MSP60K \cite{jin2025pedestrian}, and MEBOW \cite{wu2020mebow}. The modified YOLOX head structure is illustrated in Figure \ref{fig:yolox}. Through this design, the detector evolves into a feature-enriched module that simultaneously provides bounding boxes and semantically meaningful attributes, forming the foundation of FeatureSORT.

\subsection{Feature-based modules}
 \label{subsec:feature-modules}
To complement the ReID embeddings, we design additional feature modules that leverage the outputs of our feature-enriched detector. Each module captures a specific attribute of pedestrian appearance or motion, providing richer cues for data association and improving robustness in challenging scenarios such as occlusion or imperfect detections. In the following, we describe the design and role of each feature module and explain how they are integrated into the tracking framework.

\textbf{Clothes Color.} \hspace{0.3cm}
For clothing color recognition, we leverage the UPAR dataset \cite{specker2023upar}, which provides large-scale annotations of pedestrian upper- and lower-body colors. A color classification head is added to the detector, trained in a multi-label setting where each pedestrian can have different colors for upper and lower body clothing. The training objective is a binary cross-entropy loss over all color categories:

\begin{equation}
\label{eq:color_loss}
\mathcal{L}_{\text{color}} = - \sum_{k=1}^{K_{\text{col}}} \Big( y^{\text{col}}_k \log \hat{y}^{\text{col}}_k 
+ (1 - y^{\text{col}}_k) \log (1 - \hat{y}^{\text{col}}_k) \Big),
\end{equation}

where $K_{\text{col}}$ is the number of color categories, 
$y^{\text{col}}_k \in \{0,1\}$ indicates whether color $k$ is present in the ground-truth label, 
and $\hat{y}^{\text{col}}_k \in [0,1]$ is the predicted probability for that category.
During inference, we maintain a color feature bank $\mathcal{C}^{(i)}$ for each tracklet $i$, storing the most recent predictions of the color head associated with that tracklet. In each time step, for the detection assigned to tracklet $i$, we append the color prediction vector  
$
\hat{\mathbf{y}}^{\text{col}} = 
[\hat{y}^{\text{col}}_1, \dots, \hat{y}^{\text{col}}_{K_{\text{col}}}]^\top \in [0,1]^{K_{\text{col}}}
$
to the bank $\mathcal{C}^{(i)}$ as  
$
\mathcal{C}^{(i)} \leftarrow \text{append}\big(\mathcal{C}^{(i)}, \hat{\mathbf{y}}^{\text{col}}\big).
$

This temporal aggregation stabilizes the representation against noise and short-term misclassifications, allowing color features to contribute more reliably to the overall association process.

\textbf{Clothes Style.} \hspace{0.3cm}  
We treat clothes style as a multi-class classification problem. For training, we use the MSP60K dataset \cite{jin2025pedestrian}, which provides large-scale annotations of pedestrian clothing styles. The style head is trained using a binary cross-entropy loss over the top $K_{\text{sty}}$ style categories:  

\begin{equation}
\label{eq:style_loss}
\mathcal{L}_{\text{style}} = - \sum_{k=1}^{K_{\text{sty}}} \Big( y^{\text{sty}}_k \log \hat{y}^{\text{sty}}_k 
+ (1 - y^{\text{sty}}_k) \log (1 - \hat{y}^{\text{sty}}_k) \Big),
\end{equation}

where $y^{\text{sty}}_k \in \{0,1\}$ indicates whether style $k$ is present in the ground truth, and $\hat{y}^{\text{sty}}_k \in [0,1]$ is the predicted probability.  
When conducting inference, we keep a style feature bank $\mathcal{S}^{(i)}$ for each tracklet $i$, storing the most recent predictions of the style head associated to that tracklet. For the detection assigned to tracklet $i$, we append the style prediction vector  
$
\hat{\mathbf{y}}^{\text{sty}} = 
[\hat{y}^{\text{sty}}_1, \dots, \hat{y}^{\text{sty}}_{K_{\text{sty}}}]^\top \in [0,1]^{K_{\text{sty}}}
$
to the bank $\mathcal{S}^{(i)}$:  
$
\mathcal{S}^{(i)} \leftarrow \text{append}\big(\mathcal{S}^{(i)}, \hat{\mathbf{y}}^{\text{sty}}\big).
$

Specialized mentioned aggregation approach stabilizes the representation of clothing style and makes it more robust against noise and short-term misclassifications, improving its utility in the association process.

\textbf{Direction.} \hspace{0.3cm}  
We model body orientation using the MEBOW dataset \cite{wu2020mebow}, which provides rich annotations of human direction across diverse poses, lighting conditions, occlusions, and backgrounds. In this dataset, orientation is defined over $360^\circ$ and discretized into $K_{\text{dir}}=72$ bins, each covering $5^\circ$. The direction head outputs a probability vector  
$
\hat{\mathbf{y}}^{\text{dir}} = [\hat{y}^{\text{dir}}_1, \dots, \hat{y}^{\text{dir}}_{K_{\text{dir}}}]^\top \in [0,1]^{K_{\text{dir}}},
$  
where $\hat{y}^{\text{dir}}_k$ denotes the predicted probability that the orientation lies within bin $k$.  

Following \cite{wu2020mebow}, we adopt a circular Gaussian regression loss, inspired by heatmap regression in keypoint estimation:  

\begin{equation}
\label{eq:direction_loss}
\mathcal{L}_{\text{dir}} = \sum_{k=1}^{K_{\text{dir}}} \Big( \hat{y}^{\text{dir}}_k - \phi(y^{\text{dir}}, k, \sigma) \Big)^2,
\end{equation}

where $y^{\text{dir}} \in \{1, \dots, K_{\text{dir}}\}$ is the ground-truth orientation bin, and $\phi(y^{\text{dir}}, k, \sigma)$ is the circular Gaussian centered at $y^{\text{dir}}$:  

\begin{equation}
\label{eq:circ_gaussian}
\phi(y^{\text{dir}}, k, \sigma) = \frac{1}{\sqrt{2 \pi \sigma ^ 2 }} e^{ - \frac{1}{2 \sigma ^2}  min( |k - y^{\text{dir}}|, K_{\text{dir}} - |k - y^{\text{dir}}| )^2 }
\end{equation}

During inference, we maintain a direction feature bank $\mathcal{D}^{(i)}$ of size one for each tracklet $i$, storing only the most recent orientation prediction. The stored direction is obtained as the maximum-likelihood bin:  
$
\mathcal{D}^{(i)} \leftarrow \arg\max_k \hat{y}^{\text{dir}}_k.
$

\textbf{ReID Features.} \hspace{0.3cm}  
The original DeepSORT employs a lightweight CNN-based ReID network to extract appearance embeddings. In FeatureSORT, we adopt the stronger BoT ReID model \cite{luo2019strong} as the appearance feature extractor. To update the tracklet embeddings over time, we replace the gallery-based feature accumulation used in DeepSORT with an exponential moving average (EMA) strategy \cite{wang2020towards}. While the gallery approach preserves long-term information, it is sensitive to detection noise \cite{du2021giaotracker}. In contrast, the EMA method softly updates the current appearance state by combining it with new detections, which provides greater robustness against noisy inputs.  
To further stabilize the representation, we maintain a feature bank $\mathcal{E}^{(i)}$ for each tracklet $i$, which stores the most recent ReID embeddings after EMA updates. This mechanism ensures that the representation evolves smoothly and remains discriminative for reliable association. Empirically, the EMA-based updating strategy improves both matching quality and computational efficiency compared to the original approach.

\subsection{Online Tracking Algorithm}
With the outputs of the feature-enriched detector and the designed feature modules, FeatureSORT performs online multi-object tracking by combining motion prediction and feature-based association. The tracking pipeline consists of three key components. First, a noise-suppressed adaptive (NSA) Kalman filter is used to predict tracklet states more robustly than the standard Kalman filter. Second, a combined distance metric integrates motion, IoU, and the proposed feature-based similarities into a unified association cost. Finally, a matching strategy assigns detections to existing tracklets based on this cost, ensuring stable identities and reduced switches.

\textbf{NSA Kalman.} \hspace{0.3cm}  
Motion prediction is a key component of the online tracking framework. However, the standard Kalman filter is not robust to imperfect detections \cite{stadler2022modelling} and assumes a fixed measurement noise, which limits its adaptability in practice. To address this issue, we adopt the NSA Kalman filter proposed in \cite{du2021giaotracker}. This approach adjusts the measurement noise covariance based on the detection confidence score, assigning higher weights to reliable detections and lower weights to uncertain ones. As a result, the NSA Kalman filter improves robustness against detection noise and enhances the overall state update process.

 \begin{figure*}[t]
    \begin{equation}
      \label{eq:combined_dist}
      \mathbf{D}_{\text{combined}} = 
      \begin{cases}
        \lambda_{\text{mot}} \mathbf{D}_{\text{motion}} + 
        \lambda_{\text{ReID}} \mathbf{D}_{\text{ReID}} + 
        \lambda_{\text{col}} \mathbf{D}_{\text{col}} + 
        \lambda_{\text{sty}} \mathbf{D}_{\text{sty}}, 
        & \text{if IoU $> \text{IoU}_{\min}$ and $\mathbf{D}_{\text{dir}} < \text{Dir}_{\max}$}, \\[6pt]
        d_{\max} + \epsilon, & \text{otherwise}.
      \end{cases}
    \end{equation}
    \hrule
\end{figure*}

\textbf{Combined Distance.} \hspace{0.3cm}  
During inference, FeatureSORT computes a joint association cost that combines motion information with feature-based distances. For each detection–tracklet pair \((i,j)\), the feature-based distances are obtained directly from the training objectives defined in the preceding subsections:  

\begin{enumerate}[label=(\roman*)]
    \item \(\mathbf{D}_{\text{ReID}}(i,j)\): obtained from the cosine embedding distance (Eq.~\ref{eq:cosine_dist}), by comparing the embedding of detection \(j\) with those stored in the ReID feature bank \(\mathcal{E}^{(i)}\), and selecting the minimum distance.  

    \item \(\mathbf{D}_{\text{col}}(i,j)\): derived from the color classification loss (Eq.~\ref{eq:color_loss}), computed between the color prediction vector of detection \(j\) and the stored vectors in the color bank \(\mathcal{C}^{(i)}\). The minimum value is taken as the distance.  

    \item \(\mathbf{D}_{\text{sty}}(i,j)\): defined analogously from the style classification loss (Eq.~\ref{eq:style_loss}), comparing the prediction of detection \(j\) with the banked vectors in \(\mathcal{S}^{(i)}\).  

    \item \(\mathbf{D}_{\text{dir}}(i,j)\): obtained from the orientation regression loss (Eq.~\ref{eq:direction_loss}), by evaluating the prediction of detection \(j\) against the stored direction state \(\mathcal{D}^{(i)}\).  
\end{enumerate}

To form the final cost matrix for association, we combine motion, ReID, and the proposed feature-based distances in Eq.~\ref{eq:combined_dist}. Specifically, we employ the Euclidean-based motion distance $\mathbf{D}_{\text{motion}}$ and the cosine embedding distance $\mathbf{D}_{\text{ReID}}$ as in Eq.~\ref{eq:cosine_dist} \cite{aharon2022bot}, together with $\mathbf{D}_{\text{col}}, \mathbf{D}_{\text{sty}},$ and $\mathbf{D}_{\text{dir}}$ defined earlier. Following \cite{stadler2023improved}, we integrate these cues into a single combined distance $\mathbf{D}_{\text{combined}}$ using a weighted sum of motion, ReID, color, and style terms, with gating conditions imposed on IoU and direction to suppress implausible matches.

In Eq.~\ref{eq:combined_dist}, $\lambda_{\text{mot}}, \lambda_{\text{ReID}}, \lambda_{\text{col}}, \lambda_{\text{sty}}$ are weighting coefficients, $\text{IoU}_{\min}$ is a minimum overlap threshold, $\text{Dir}_{\max}$ is a maximum admissible direction distance, $d_{\max}$ is a large penalty value, and $\epsilon$ is a small constant for numerical stability.
This formulation encourages associations only when both spatial overlap and directional consistency are satisfied, while penalizing implausible matches. Compared to a simple unweighted sum of costs, this combined distance improves robustness by leveraging the complementary strengths of motion, ReID, and the newly introduced feature modules.

\textbf{Matching.} \hspace{0.3cm} For association, we employ the Hungarian algorithm, which provides an exact solution to the global linear assignment problem. Unlike the matching cascade strategy used in DeepSORT, which introduces hand-crafted prioritization rules, the Hungarian method directly optimizes the assignment over the entire cost matrix. Prior work \cite{zhang2022bytetrack} has shown that as trackers become stronger and more robust to ambiguous associations, additional heuristic constraints may actually degrade performance. Consistent with this observation, we replace the cascade matching with the Hungarian algorithm to achieve more accurate and globally optimal associations.

\subsection{Post Processing}
Beyond online tracking, we adopt lightweight post-processing methods to refine trajectories. A Gaussian smoothing process (GSP) is applied to reduce noise and interpolate missing detections, while a global linking algorithm merges fragmented tracklets into longer, more consistent trajectories. These steps are model-independent and serve as optional enhancements that further improve accuracy without altering the online tracking pipeline.

\textbf{GSP.} \hspace{0.3cm}  
After updating tracklets, we apply a GSP to refine trajectories and handle missing detections. Unlike linear interpolation, which simply connects gaps without considering motion dynamics, GSP models trajectories as a Gaussian process \cite{schulz2018tutorial}. This allows each missing position to be corrected using both past and future observations, resulting in smoother and more realistic trajectories. The method is lightweight, model-independent, and improves robustness against occlusion and noisy detections without altering the online tracking pipeline.

\textbf{Global Linking.} \hspace{0.3cm}  
To merge fragmented trajectories, we adopt the global linking algorithm from GIAOTracker \cite{du2021giaotracker}. This method learns spatio-temporal embeddings of tracklets using an improved ResNet50-TP backbone with temporal modeling and part-level supervision, which improves robustness to occlusion and appearance changes. Tracklets are then associated globally using Hungarian matching based on embedding similarity and spatiotemporal consistency. By incorporating this global reasoning step, fragmented tracklets caused by missed detections or occlusion can be reconnected into longer, more reliable trajectories.

\section{Results}
\label{sec:results}
In this section, we first describe the experimental setup, then present ablation studies on individual modules, and finally report benchmark evaluations.

\subsection{Experimental Setup}

\textbf{Datasets.} \hspace{0.3cm}  
We conduct experiments on the MOT16, MOT17 \cite{zhou2020tracking}, MOT20 \cite{dendorfer2020mot20} and DanceTrack \cite{sun2022dancetrack} benchmarks. MOT17 contains 7 training sequences (5,316 frames) and 7 test sequences (5,919 frames), while MOT20 consists of 4 training sequences (8,931 frames) and 4 test sequences (4,479 frames), featuring highly crowded and challenging pedestrian scenes.
DanceTrack \cite{sun2022dancetrack} is a recent benchmark with frequent occlusions and non-rigid motions, designed to test MOT models under more challenging conditions.
Following prior works \cite{ du2023strongsort}, we use the first half of each training sequence for training and the second half for validation. For ablation studies, the detector is trained using half of the MOT16, MOT17, MOT20 and DanceTrack training data, as in \cite{wu2021track}.

\textbf{Metrics.} \hspace{0.3cm}  
Tracking performance is assessed using a combination of established metrics. We report the CLEAR MOT metrics \cite{bernardin2008evaluating}—including MOTA, false positives (FP), false negatives (FN), and identity switches (IDs)—together with IDF1 \cite{ristani2016performance} and the more recent HOTA metric \cite{luiten2021hota}. MOTA provides an overall measure of detection and association quality, IDF1 emphasizes the correctness of identity preservation, and HOTA offers a balanced evaluation that jointly accounts for detection, association, and localization accuracy.

\textbf{Implementation Details.} \hspace{0.3cm}  
Our training schedule and detector settings (classification and localization heads) follow the optimal configurations reported in prior works \cite{zhang2022bytetrack, du2023strongsort}. For the feature heads (color, style, and direction), we freeze the rest of the network and train only the added heads for 50 additional epochs, using a scheduled learning rate with exponential decay tuned by grid search. During inference, the non-maximum suppression (NMS) threshold is set to 0.8 and the detection confidence threshold to 0.5. For matching, we use $\text{IoU}_{\min}=0.45$, while the momentum term in EMA is $\alpha=0.8$. The distance weighting coefficients are set to $\lambda_{\text{motion}}=0.3$, $\lambda_{\text{edge}}=0.4$, $\lambda_{\text{color}}=0.25$, and $\lambda_{\text{style}}=0.25$, with the maximum admissible direction distance fixed at $\text{Dir}_{\max}=0.85$.

Deleting tracklet age is set to 10 frames which are not updated for the periods more than this threshold.
The maximum gap allowed for interpolation in GSP is 20 frames. In addition to the ReID dataset. The input size of the global linking network is set to [224, 112, 4] for pedestrian tracklets. We utilize Adam as the optimizer and cross-entropy loss as the objective function for training the global linking network over 50 epochs, employing a learning rate schedule of cosine annealing. During inference, we apply a temporal distance threshold of 20 frames and a spatial distance threshold of 70 pixels to remove unrealistic association pairs. Finally, the association is
accepted if its score predicted by the network is larger than 0.9.
All experiments are conducted on an NVIDIA RTX 4060 8GB GPU.

\subsection{Ablation Studies}
We divide our ablation studies into three parts. First, we examine how our feature-based modules introduced in section \ref{subsec:feature-modules} affect overall tracking performance. Second, we compare our proposed vanilla matching with the cascade matching in DeepSORT \cite{wojke2017simple} to show the effectiveness of the suggested matching algorithm. Finally we demonstrate the necessity of post-processing by evaluating GSP and Global Linking.

\textbf{Ablation study for feature-based modules.} \hspace{0.3cm}
Table~\ref{tab:ablation1} details the incremental enhancements applied to DeepSORT to form FeatureSORT and empirically demonstrates the effectiveness of each module.

\begin{table*}[t]
    \caption{
            Ablation study on the MOT17 validation set evaluating the contribution of different components in FeatureSORT. 
The tested modules include: stronger ReID embedding extractor (BoT \cite{luo2019strong}), NSA Kalman filter (NSA \cite{du2021giaotracker}), EMA feature updating (EMA \cite{wang2020towards}), combined distance cost (CD), vanilla Hungarian matching without cascade (VM), and the proposed feature modules (color, style, direction). 
Results are reported in the online setting without offline post-processing. (Best in \textbf{bold}.)
}
	\begin{minipage}{1\linewidth} 
		\label{tab:ablation1}
		\centering
		\small
		        \begin{tabular}{c | c c    c c    c c    c c    c c    cc }
                    \toprule
                    \multicolumn{1}{c}{Method} & \multicolumn{1}{c}{BoT \cite{luo2019strong}} & \multicolumn{1}{c}{NSA  \cite{du2021giaotracker}} & \multicolumn{1}{c}{EMA \cite{wang2020towards}} & \multicolumn{1}{c}{CD} & \multicolumn{1}{c}{VM} & \multicolumn{1}{c}{Color} & \multicolumn{1}{c}{Style} & \multicolumn{1}{c}{Dir} & \multicolumn{1}{c}{MOTA$\uparrow$} & \multicolumn{1}{c}{HOTA$\uparrow$} & \multicolumn{1}{c}{IDF1$\uparrow$} & \multicolumn{1}{c}{IDs$\downarrow$}\\
                     \hline
                    Vanilla Model & - & - & - & - & - & - & - & - & 75.3 & 60.8 & 74.8 & 2897\\
                    FeatureSORTv1& \checkmark & - & - & - & - & - & - & - & 77 & 61 & 75.5 & 2831\\
                    FeatureSORTv2 & \checkmark & \checkmark & - & - & - & - & - & - & 77.2 & 61.1 & 75.6 & 2766\\
                    FeatureSORTv3 & \checkmark & \checkmark & \checkmark & - & - & - & - & - & 77.1 & 62.2 & 75.6 & 1698 \\
                    FeatureSORTv4 & \checkmark & \checkmark & \checkmark & \checkmark & - & - & - & - & 77.7 & 62.5 & 75.9 & 2644 \\
                    FeatureSORTv5 & \checkmark & \checkmark & \checkmark & \checkmark & \checkmark & - & - & - & 78.3 & 62.5 & 76.2 & 2692 \\
                    FeatureSORTv6 & \checkmark & \checkmark & \checkmark & \checkmark & \checkmark & \checkmark & - & - & 79.4 & 63 & 77.2 & 2340 \\
                    FeatureSORTv7 & \checkmark & \checkmark & \checkmark & \checkmark & \checkmark & \checkmark & \checkmark & - & \textbf{80.9} & \textbf{64.4} & 77 & 2583 \\
                    FeatureSORTv8 & \checkmark & \checkmark & \checkmark & \checkmark & \checkmark & - & - & \checkmark & 78.4 & 62.5 & 76.8 & \textbf{976} \\
                    FeatureSORTv9 & \checkmark & \checkmark & \checkmark & \checkmark & \checkmark & \checkmark & \checkmark & \checkmark & 79.9 & 63.2 & \textbf{77.4} & 2212\\
                    \hline
                \end{tabular}
	\end{minipage}
\end{table*}

\begin{figure*}[t]
\vspace{-0.4cm}
\centering
\includegraphics[height=5.5cm, width=18cm]{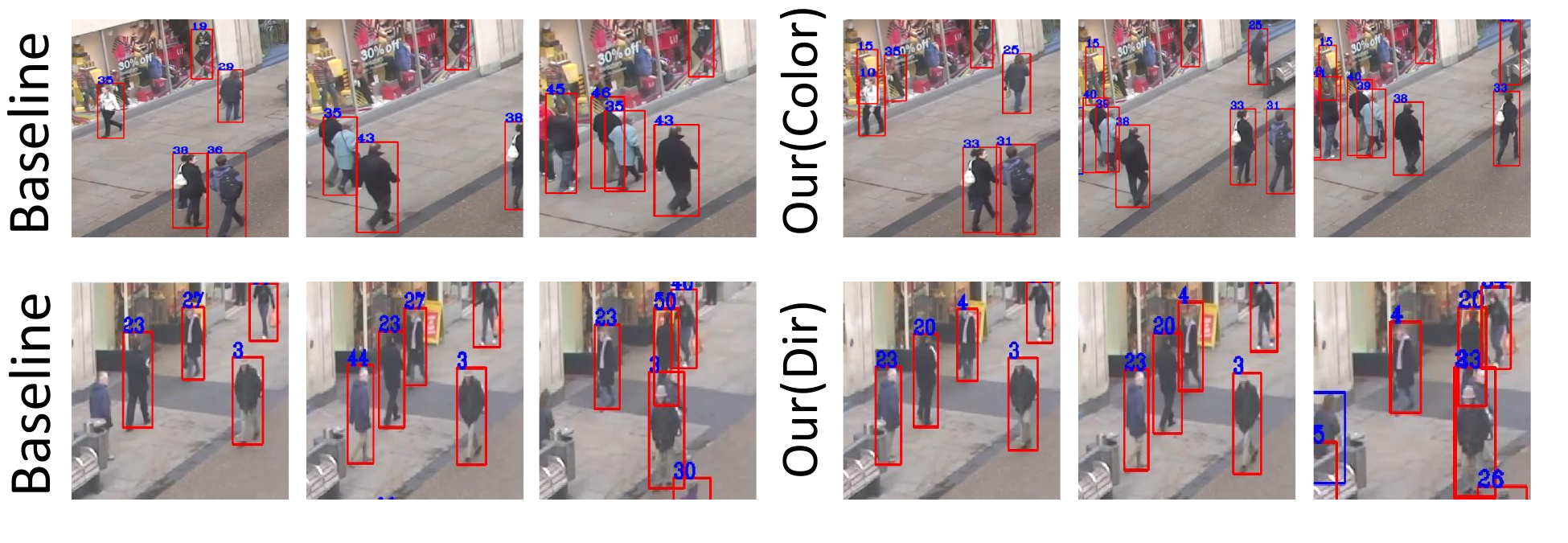}
\caption{Ablation study illustrating the impact of feature modules on ID consistency. 
\textbf{Top row:} Without using color features (left 3 frames), ID switches occur among pedestrians with similar appearance. Incorporating color features (right 3 frames) resolves the ambiguity and maintains consistent IDs. 
\textbf{Bottom row:} Without direction cues (left 3 frames), ID switches occur during crossing motions. Using direction features (right 3 frames) prevents the switches by enforcing orientation consistency.}
\setlength{\intextsep}{-10pt}
\vskip -0.05in
\label{fig:color_dir_effect}
\end{figure*}

\begin{enumerate}
    \item \textbf{BoT:} Replacing the original ReID embedding extractor with BoT improves MOTA by +1.7 and IDF1 by +0.7, highlighting the importance of stronger appearance embeddings for reliable association.
    \item \textbf{NSA:} Incorporating the NSA Kalman filter yields an additional gain of +0.2 in MOTA and +0.1 in IDF1, reflecting improved motion modeling through adaptive noise adjustment in the update step.
    \item \textbf{EMA:} The EMA feature updating mechanism improves HOTA by +1.1, stabilizing appearance embeddings over time and leading to more consistent associations.
    \item \textbf{CD:} Incorporating the combined distance (motion, edge, and IoU terms as in Eq.~\ref{eq:combined_dist}) improves MOTA by +0.6, IDF1 by +0.3, and HOTA by +0.3, showing that integrating multiple cues yields more reliable associations. (In FeatureSORTv4, only motion, edge, and IoU distances are used.)
    
    \item \textbf{VM:} Replacing cascade matching with vanilla Hungarian matching improves MOTA by +0.6 and IDF1 by +0.3, confirming that stronger trackers benefit from global assignment without additional heuristic constraints.
    \item \textbf{Color:} Adding color features to the cost function yields a clear gain of +1.1 in MOTA and +1.0 in IDF1. As illustrated in Fig.~\ref{fig:color_dir_effect} (top row), color cues help disambiguate visually similar pedestrians and prevent ID switches.  
    \item \textbf{Style:} Incorporating clothing style features into the cost function improves MOTA by +1.5 and HOTA by +1.4. However, we also observe an increase in ID switches, likely because highly detailed style cues can occasionally prevent matches and trigger new IDs for the same object.
    \item \textbf{Direction:} Incorporating directional features reduces ID switches significantly, with a notable improvement reflected in the ablation results. Fig.~\ref{fig:color_dir_effect} (bottom row) shows how orientation cues resolve mismatches during crossing.
\end{enumerate}

\textbf{Ablation study for vanilla matching.} \hspace{0.3cm}  
Table~\ref{tab:ablation3} compares cascade matching with vanilla Hungarian matching across different versions of FeatureSORT. The results show that cascade matching yields better performance for the baseline DeepSORT. However, its benefit diminishes for FeatureSORT variants and can even harm tracking accuracy in stronger models. This suggests that while the prior constraints in cascade matching help reduce ambiguous associations in weaker trackers, they impose unnecessary restrictions on stronger trackers, ultimately limiting performance \cite{du2023strongsort}.

\begin{table}[h]
    \caption{
            Ablation study on the MOT17 validation set comparing the cascade matching algorithm and vanilla Hungarian matching. 
All results are reported in the online setting without offline post-processing. Best results are highlighted in bold.
        }
	\resizebox{\linewidth}{!}{
		\label{tab:ablation3}
		\centering
		        \begin{tabular}{c | c  cccc }
                    \toprule
                    \multicolumn{1}{c}{Method} & \multicolumn{1}{c}{Matching} & \multicolumn{1}{c}{MOTA$\uparrow$} & \multicolumn{1}{c}{HOTA$\uparrow$} & \multicolumn{1}{c}{IDF1$\uparrow$} & \multicolumn{1}{c}{IDs$\downarrow$}\\
                     \hline
                     \multicolumn{1}{c}{\multirow{2}{*}{DeepSORT}} & cascade & 77.3 & 60.5 & 76 & 3374\\
                    \multicolumn{1}{c}{} & vanilla & 76.2 & 60.4 & 76 & 3416 \\
                    \hline
                    \multicolumn{1}{c}{\multirow{2}{*}{FeatureSORTv5}} & cascade & 77.8 & 62.5 & 76.1 & 2750\\
                    \multicolumn{1}{c}{} & vanilla & 78.3 & 62.5 & 76.2 & 2692 \\
                    \hline
                    \multicolumn{1}{c}{\multirow{2}{*}{FeatureSORTv6}} & cascade & 79.2 & 62.8 & 76.7 & 2402\\
                    \multicolumn{1}{c}{} & vanilla & 79.6 & 63.2 & 77.4 & 2322 \\
                    \hline
                    \multicolumn{1}{c}{\multirow{2}{*}{FeatureSORTv7}} & cascade & 80.7 & 64.2 & 76.7 & 2590 \\
                    \multicolumn{1}{c}{} & vanilla & \textbf{80.9} & \textbf{64.4} & 77 & 2583\\
                    \hline
                    \multicolumn{1}{c}{\multirow{2}{*}{FeatureSORTv8}} & cascade & 78.3 & 62.5 & 76.7 & 1062 \\
                    \multicolumn{1}{c}{} & vanilla & 78.4 & 62.5 & 76.8 & \textbf{976}  \\
                    \hline
                    \multicolumn{1}{c}{\multirow{2}{*}{FeatureSORTv9}} & cascade & 79.7 & 63.2 & 77.4 & 2260 \\
                    \multicolumn{1}{c}{} & vanilla & 79.9 & 63.2 & \textbf{77.4} & 2212 \\
                    \hline
                \end{tabular}
	}
\end{table}

\textbf{Ablation study for Global Linking and GSP.} \hspace{0.3cm}  
Table~\ref{tab:ablation2} reports the effect of Global Linking and GSP on three FeatureSORT variants and three state-of-the-art trackers (CenterTrack \cite{zhou2020tracking}, TransTrack \cite{sun2020transtrack}, and FairMOT \cite{zhang2021fairmot}). Global Linking consistently improves results, with larger gains for weaker trackers that are more vulnerable to missing associations. GSP further enhances stronger trackers by smoothing trajectories and reducing noise, but is less effective on weaker trackers where false associations dominate.

\begin{table}[h]
    \caption{
            Effect of GlobalLink and GSP applied to different MOT methods. 
Best results are in \textbf{bold}.
        }
	\resizebox{\linewidth}{!}{
		\label{tab:ablation2}
		\centering
		        \begin{tabular}{c | c c    c c    c }
                    \toprule
                    \multicolumn{1}{c}{Method} & \multicolumn{1}{c}{GlobalLink} & \multicolumn{1}{c}{GSP} & \multicolumn{1}{c}{MOTA$\uparrow$} & \multicolumn{1}{c}{HOTA$\uparrow$} & \multicolumn{1}{c}{IDF1$\uparrow$} \\
                     \hline
                    \multicolumn{1}{c}{\multirow{3}{*}{FeatureSORTv5}} & - & - & 78.3 & 62.5 & 76.2 \\
                    \multicolumn{1}{c}{} & \checkmark & - & 80.5 & 62.6 & 76.3 \\
                    \multicolumn{1}{c}{} & \checkmark & \checkmark & \textbf{80.6} & \textbf{62.9} & \textbf{76.4} \\
                    \hline
                    \multicolumn{1}{c}{\multirow{3}{*}{FeatureSORTv7}} & - & - & 80.9 & 64.4 & 77 \\
                    \multicolumn{1}{c}{} & \checkmark & - & 81 & 64.9 & 77.2 \\
                    \multicolumn{1}{c}{} & \checkmark & \checkmark & \textbf{81.5} & \textbf{65.1} & \textbf{77.2} \\
                    \hline
                    \multicolumn{1}{c}{\multirow{3}{*}{FeatureSORTv9}} & - & - & 79.9 & 63.2 & 77.4 \\
                    \multicolumn{1}{c}{} & \checkmark & - & 80.2 & 63.5 & 77.6 \\
                    \multicolumn{1}{c}{} & \checkmark & \checkmark & \textbf{80.5} & \textbf{63.4} & \textbf{77.8} \\
                    \hline
                    \multicolumn{1}{c}{\multirow{3}{*}{CenterTrack \cite{zhou2020tracking}}} & - & - & 67.8 & 60.3 & 64.7 \\
                    \multicolumn{1}{c}{} & \checkmark & - & 69.4 & 61 & 65 \\
                    \multicolumn{1}{c}{} & \checkmark & \checkmark & \textbf{69.7} & \textbf{61.1} & \textbf{65.3} \\
                    \hline
                    \multicolumn{1}{c}{\multirow{3}{*}{TransTrack \cite{sun2020transtrack}}} & - & - & 74.5 & 54.1 & 63.9 \\
                    \multicolumn{1}{c}{} & \checkmark & - & 74.8 & 54.5 & 64 \\
                    \multicolumn{1}{c}{} & \checkmark & \checkmark & \textbf{75} & \textbf{54.8} & \textbf{64.3} \\
                    \hline
                    \multicolumn{1}{c}{\multirow{3}{*}{FairMOT \cite{zhang2021fairmot}}} & - & - & 73.7 & 59.3 & 72.3 \\
                    \multicolumn{1}{c}{} & \checkmark & - & 74.1 & \textbf{59.6} & \textbf{72.3} \\
                    \multicolumn{1}{c}{} & \checkmark & \checkmark & \textbf{74.4} & 59.6 & 72.2 \\
                    \hline
                \end{tabular}
	}
\end{table}

\textbf{Tracking performance with lightweight YOLOX.} \hspace{0.3cm}  
We compare FeatureSORT and DeepSORT using progressively smaller YOLOX variants as the detector backbone. FeatureSORT employs our modified YOLOX with four heads (including feature heads), whereas DeepSORT relies on the standard two-head YOLOX. Both detectors are trained on the same bounding box datasets, with additional pedestrian attribute datasets used to train the feature heads of FeatureSORT. The results in Table~\ref{tab:light_models} show that FeatureSORT consistently achieves higher MOTA, HOTA, and IDF1 across all detector sizes. Notably, with the lightweight YOLOX-Nano, FeatureSORT attains a MOTA that is 4.7 points higher than DeepSORT, underscoring its suitability for real-world, resource-constrained applications.

\begin{table}[h]
    \caption{Comparison of FeatureSORT and DeepSORT with lightweight YOLOX detectors on the MOT17 validation set.}
	\resizebox{\linewidth}{!}{ 
		\label{tab:light_models}
		\centering
		        \begin{tabular}{c  c  c c c c}
                    \toprule
                    \multicolumn{1}{c}{Detector} & \multicolumn{1}{c}{Params} & \multicolumn{1}{c}{Tracker} & \multicolumn{1}{c}{MOTA$\uparrow$} & \multicolumn{1}{c}{HOTA$\uparrow$} & \multicolumn{1}{c}{IDF1$\uparrow$} \\
                     \hline
                    \multirow{2}{*}{YOLOX-L} & \multirow{2}{*}{61M} & FeatureSORTv9 & 77.8 & 62 & 76.6 \\
                    \multicolumn{1}{c}{} & \multicolumn{1}{c}{} & DeepSORT & 75.6 & 59.9 & 73 \\
                    \hline
                    \multirow{2}{*}{YOLOX-M} & \multirow{2}{*}{28M} & FeatureSORTv9 & 76.5 & 60.6 & 74.9 \\
                    \multicolumn{1}{c}{} & \multicolumn{1}{c}{} & DeepSORT & 74.5 & 59.1 & 76.9 \\
                    \hline
                    \multirow{2}{*}{YOLOX-S} & \multirow{2}{*}{10.5M} & FeatureSORTv9 & 72.9 & 58.9 & 74.2 \\
                    \multicolumn{1}{c}{} & \multicolumn{1}{c}{} & DeepSORT & 69.6 & 56 & 71.5 \\
                    \hline
                    \multirow{2}{*}{YOLOX-Tiny} & \multirow{2}{*}{5.8M} & FeatureSORTv9 & 72.6 & 58.1 & 73.5 \\
                    \multicolumn{1}{c}{} & \multicolumn{1}{c}{} & DeepSORT & 68.6 & 54.6 & 72 \\
                    \hline
                    \multirow{2}{*}{YOLOX-Nano} & \multirow{2}{*}{1M} & FeatureSORTv9 & 66.1 & 52.5 & 69.4 \\
                    \multicolumn{1}{c}{} & \multicolumn{1}{c}{} & DeepSORT & 61.4 & 49 & 66.8 \\
                    \hline
                \end{tabular}
	}
\end{table}

\subsection{Benchmark Evaluation}
We evaluate FeatureSORT on the MOT16, MOT17, MOT20 and testsets and compare it against recent SOTA online MOT trackers, using the official benchmark metrics.

\textbf{MOT16 and MOT17.} \hspace{0.2cm}
The performance of FeatureSORT in online mode without post-processing on the MOT16 and MOT17 test sets is reported in Tables~\ref{tab:mot16} and~\ref{tab:mot17}. 
On MOT16, FeatureSORT achieves higher MOTA than SGT~\cite{hyun2205detection}, FairMOT~\cite{zhang2021fairmot}, and StrongSORT~\cite{du2023strongsort} by roughly 3\%, 5\%, and 2\%, respectively. 
On MOT17, the improvements over these methods are about 4\%, 7\%, and 2\%, with the note that FairMOT is trained with additional datasets. 
Without the BoT ReID network, FeatureSORT can be regarded as a joint detector–tracker, making it directly comparable to FairMOT. 
FeatureSORTv9 achieves strong MOTA, HOTA, and IDF1 on both MOT16 and MOT17 through a balanced trade-off between FP, FN, and IDs.  
Compared to methods using the same detector, such as StrongSORT and ByteTrack~\cite{zhang2022bytetrack}, FeatureSORT yields consistent gains across most MOT metrics. 
It also surpasses OMC-F~\cite{liang2022one}, designed for low-score detections, by achieving higher MOTA, HOTA, and IDF1, while reducing both false positives and false negatives.

\begin{table}[h]
    \caption{
            MOT16: Performance comparison with SOTA
        }
	\resizebox{\linewidth}{!}{ 
		\label{tab:mot16}
		\centering
		        \begin{tabular}{| c | c  ccccc }
                    \toprule
                    \multicolumn{1}{c}{Method} & \multicolumn{1}{c}{MOTA$\uparrow$} & \multicolumn{1}{c}{HOTA$\uparrow$} & \multicolumn{1}{c}{IDF1$\uparrow$} & \multicolumn{1}{c}{FP$\downarrow$} & \multicolumn{1}{c}{FN$\downarrow$} & \multicolumn{1}{c}{IDs$\downarrow$} \\
                    \hline
                    \multicolumn{1}{c}{QDTrack \cite{pang2021quasi}} & 69.8  & 54.5 & 67.1 	 & 9861& 44050& 1097\\
                    \multicolumn{1}{c}{TraDes \cite{wu2021track}} & 70.1 & 53.2 & 64.7 & \textbf{8091} & 45210& 1144 \\
                    \multicolumn{1}{c}{CSTrack \cite{liang2022rethinking}} & 75.6  & 59.8 & 73.3  & 9646& 33777& 1121\\
                    \multicolumn{1}{c}{GSDT \cite{wang2021joint}} & 74.5 & 56.6 & 68.1 & 8913& 36428& 1229 \\
                    \multicolumn{1}{c}{RelationTrack \cite{yu2022relationtrack}} & 75.6 & 61.7 & 75.8 & 9786& 34214& \textbf{448} \\
                    \multicolumn{1}{c}{OMC \cite{liang2022one}} & 76.4 & 62
                    9& 74.1 & 10821 & 31044& 1087\\
                    \multicolumn{1}{c}{CorrTracker \cite{wang2021multiple}} & 76.6  & 61.0 & 74.3 & 10860& 30756& 979 \\
                    \multicolumn{1}{c}{SGT \cite{hyun2205detection}} & 76.8  & 61.2 & 73.5 & 10695 & 30394& 1276\\
                    \multicolumn{1}{c}{FairMOT \cite{zhang2021fairmot}} & 74.9 & 58.3 & 72.8 & 9952& 38451& 1074\\
                    \multicolumn{1}{c}{StrongSORT \cite{du2023strongsort}} & 77.8 & 63.8 & \textbf{78.3} & 11254 & 32584 & 1538\\
                    \hline
                    \hline
                    \rowcolor{gray!20}
                    \multicolumn{1}{c}{FeatureSORTv7} & \textbf{79.7}  & \textbf{63.8} & 76.6 & 9182 & 27563 & 1658\\
                    \rowcolor{gray!20}
                    \multicolumn{1}{c}{FeatureSORTv8} & 77.9  & 62.8 & 76.3 & 14827 & 26877 & 597\\
                    \rowcolor{gray!20}
                    \multicolumn{1}{c}{FeatureSORTv9} & 78.3  & 63.1 & 76.8 & 13321 & \textbf{24992} & 1336\\
                    \hline
                \end{tabular}
	}
\end{table}

\begin{table}[h]
    \caption{
            MOT17: Performance comparison with SOTA
        } 
		\label{tab:mot17}
		\centering
		\resizebox{\linewidth}{!}{
		        \begin{tabular}{| c | c  ccccc }
                    \toprule
                    \multicolumn{1}{c}{Method} & \multicolumn{1}{c}{MOTA$\uparrow$} & \multicolumn{1}{c}{HOTA$\uparrow$} & \multicolumn{1}{c}{IDF1$\uparrow$} & \multicolumn{1}{c}{FP$\downarrow$} & \multicolumn{1}{c}{FN$\downarrow$} & \multicolumn{1}{c}{IDs$\downarrow$} \\
                    \hline
                    \multicolumn{1}{c}{CTracker \cite{peng2020chained}} & 66.6 & 49.0  & 57.4  & 22284 & 160491 & 5529\\
                    \multicolumn{1}{c}{CenterTrack \cite{zhou2020tracking}} & 67.8 & 60.3 & 64.7 & \textbf{18489} & 160332 & 3039 \\
                    \multicolumn{1}{c}{QDTrack \cite{pang2021quasi}} & 68.7 & 63.5 & 66.3 & 26598 & 146643 & 3378\\
                    \multicolumn{1}{c}{TraDes \cite{wu2021track} } & 69.1 & 52.7  & 63.9 & 20892 & 150060 & 3555 \\
                    \multicolumn{1}{c}{SOTMOT \cite{zheng2021improving} } & 71 & 64.1 & 71.9 & 39537 & 118983 & 5184 \\
                    \multicolumn{1}{c}{GSDT \cite{wang2021joint} } & 73.2 & 55.2 & 66.5 & 26397 & 120666 & 3891\\
                    \multicolumn{1}{c}{RelationTrack \cite{yu2022relationtrack}} & 73.8 & 59.9 & 74.7 & 27999 & 118623 & 1374 \\
                    \multicolumn{1}{c}{TransTrack \cite{sun2020transtrack}} & 74.5 & 54.1 & 63.9 & 28323 & 112137 & 3663 \\
                    \multicolumn{1}{c}{OMC–F \cite{liang2022one}} & 74.7 & 56.8 & 73.8 & 30162 & 108556 & -\\
                    \multicolumn{1}{c}{CSTrack \cite{liang2022rethinking}} & 74.9 & 59.3 & 72.3 & 23847 & 114303 & 3567\\
                    \multicolumn{1}{c}{OMC \cite{liang2022one}} & 76 & 57.1 & 73.8 & 28894 & 101022 & -\\
                    \multicolumn{1}{c}{SGT \cite{hyun2205detection}} & 76.3 & 57.3 & 72.8 & 25974 & 102885 & 4101\\
                    \multicolumn{1}{c}{CorrTracker \cite{wang2021multiple}} & 76.4 & 58.4 & 73.6 & 29808 & 99510 & 3369 \\
                    \multicolumn{1}{c}{FairMOT \cite{zhang2021fairmot} } & 73.7 & 59.3  & 72.3 & 27507 & 117477 & 3303\\
                    \multicolumn{1}{c}{DeepSORT \cite{wojke2017simple} } & 78 & 61.2  & 74.5 & 29852 & 94716 & 1821\\
                    \multicolumn{1}{c}{ByteTrack \cite{zhang2022bytetrack}} & 78.9 & 62.8 & 77.2 & 25491 & 83721 & 2196 \\
                    \multicolumn{1}{c}{StrongSORT \cite{du2023strongsort}} & 78.3 & 63.5 & \textbf{78.5} & 27876 & 86205 & 1446 \\
                    \hline
                    \hline
                    \rowcolor{gray!20}
                    \multicolumn{1}{c}{FeatureSORTv7} & \textbf{80.6}  & \textbf{64.2} & 76.7 & 27581 & 84362 & 2637\\
                    \rowcolor{gray!20}
                    \multicolumn{1}{c}{FeatureSORTv8} & 78.4  & 62.4 & 76.6 & 30294 & 87928 & \textbf{1023}\\
                    \rowcolor{gray!20}
                    \multicolumn{1}{c}{FeatureSORTv9} & 79.6  & 63 & 77.2 & 29588 & \textbf{83132} & 2269\\
                    \hline
                \end{tabular}
	}
\end{table}

\begin{figure}[h]
    \centering
    \includegraphics[width=0.9\linewidth, keepaspectratio]{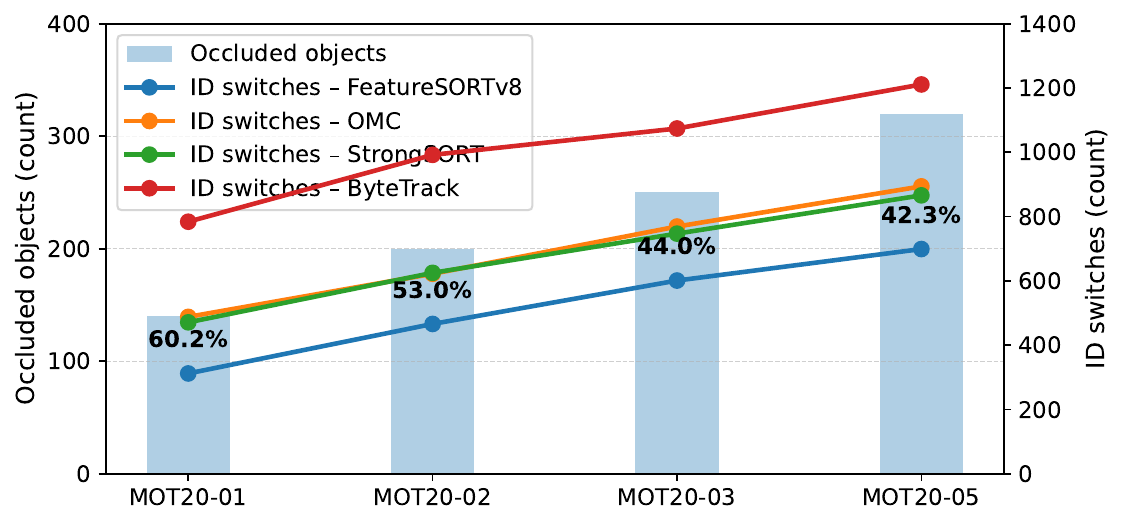}
    \caption{Reduction in ID switches vs.\ baseline. Bars: occluded objects (left axis). Lines: ID switches per method (right axis).
   }
    \label{fig:occlusion_idswitch}
\end{figure}

\textbf{MOT20.} \hspace{0.3cm}
MOT20, released after MOT16–17, focuses on highly crowded scenes with frequent partial occlusions. 
With the same detection threshold, existing methods often suffer from missed detections. 
Some works~\cite{zhang2021fairmot, wang2021multiple} lower the threshold to recover these detections, but this leads to more false positives, ID switches, and lower IDF1, as their pairwise relational features struggle to handle the large number of candidates.  
In contrast, FeatureSORT leverages feature-based modules and higher-order relational cues, achieving SOTA performance on MOT20 (Table~\ref{tab:mot20}). 
FeatureSORTv7 surpasses CorrTracker~\cite{wang2021multiple} by 12\% and SGT~\cite{hyun2205detection} by 13\% in MOTA, while FeatureSORTv9 offers a better balance between FN, FP, and IDs, outperforming OMC, StrongSORT, and ByteTrack across MOTA and HOTA. 
Although OMC uses past frames to select low-confidence detections, its reliance on pairwise matching limits its effectiveness; FeatureSORT’s vanilla matching avoids these constraints and achieves superior results.  
Finally, FeatureSORTv8 highlights the effectiveness of directional cues in crowded scenes: by enforcing orientation consistency, it markedly reduces ID switches during pedestrian crossings.
Figure~\ref{fig:occlusion_idswitch} shows the relationship between occlusion levels in four MOT20 sequences and the number of ID switches. The percentages in the bars indicate the relative reduction in ID switches of FeatureSORTv8 compared to the baseline ByteTrack.

\begin{table}[h]
    \caption{
            MOT20: Performance comparison with SOTA
        }
	\resizebox{\linewidth}{!}{
		\label{tab:mot20}
		\centering
		        \begin{tabular}{| c | c  ccccc }
                    \toprule
                    \multicolumn{1}{c}{Method} & \multicolumn{1}{c}{MOTA$\uparrow$} & \multicolumn{1}{c}{HOTA$\uparrow$} & \multicolumn{1}{c}{IDF1$\uparrow$} & \multicolumn{1}{c}{FP$\downarrow$} & \multicolumn{1}{c}{FN$\downarrow$} & \multicolumn{1}{c}{IDs$\downarrow$} \\
                    \hline
                    \multicolumn{1}{c}{SORT \cite{bewley2016simple}} & 42.7 & 36.1 & 45.1 & 28398 & 287582 & 4852 \\
                    \multicolumn{1}{c}{Tracktor++ \cite{bergmann2019tracking} } & 52.6 & 42.1 & 52.7 & 35536 & 236680 &  1648\\
                    \multicolumn{1}{c}{CSTrack \cite{liang2022rethinking}  } & 66.6 & 54 & 68.6 & 25404 & 144358 & 3196 \\
                    \multicolumn{1}{c}{CrowdTrack \cite{stadler2021performance} } & 70.7 & 55 & 68.2 & 21928 & 126533 & 3198 \\
                    \multicolumn{1}{c}{RelationTrack \cite{yu2022relationtrack}} & 67.2 & 56.5 & 70.5 & 61134 & 104597 & 4243\\
                    \multicolumn{1}{c}{DeepSORT \cite{wojke2017simple}} & 71.8 & 57.1 & 69.6 & 37858 & 101581 & 3754 \\
                    \multicolumn{1}{c}{TransTrack \cite{sun2020transtrack} } & 64.5 & 48.9 & 59.2 & 28566 & 151377 & 3565\\
                    \multicolumn{1}{c}{CorrTracker \cite{wang2021multiple}} & 65.2 & 57.1 & 69.1 & 79429 & 95855 & 5193\\
                    \multicolumn{1}{c}{GSDT \cite{wang2021joint} } & 67.1 & 53.6 & 67.5 & 31913 & 135409 & 3131 \\
                    \multicolumn{1}{c}{SOTMOT \cite{zheng2021improving}} & 68.6 & 55.7 & 71.4 & 57064 & 101154 & 4209
                    \\
                    \multicolumn{1}{c}{OMC \cite{liang2022one} } & 73.1 & 60.5 & 74.4 & \textbf{16159} & 108654 & 779 
                    \\
                    \multicolumn{1}{c}{FairMOT \cite{zhang2021fairmot} } & 61.8 & 54.6 & 67.3 & 103404 & 88901 & 5243
                    \\
                    \multicolumn{1}{c}{SGT \cite{hyun2205detection}} & 64.5 & 56.9 & 62.7 & 67352 & 111201 & 4909
                    \\
                    \multicolumn{1}{c}{ByteTrack \cite{zhang2022bytetrack} } & 75.7 & 60.9 & 74.9 & 26249 & \textbf{87594} & 1223
                    \\
                    \multicolumn{1}{c}{StrongSORT \cite{du2023strongsort}} & 72.2 & 61.5 & \textbf{75.9} & 16632 & 117920 & 770
                    \\
                    \hline
                    \hline
                    \rowcolor{gray!20}
                    \multicolumn{1}{c}{FeatureSORTv7} & \textbf{77.9}  & \textbf{63.4} & 75 & 26223 & 89050 & 1794\\
                    \rowcolor{gray!20}
                    \multicolumn{1}{c}{FeatureSORTv8} & 76  & 61.1 & 74.7 & 27541 & 99582 & \textbf{693}\\
                    \rowcolor{gray!20}
                    \multicolumn{1}{c}{FeatureSORTv9} & 76.6  & 61.3 & 75.1 & 25083 & 95027 & 1081\\
                    \hline
                \end{tabular}
	}
\end{table}

\textbf{DanceTrack.} \hspace{0.2 cm}
  We include comparison with the DanceTrack test set to demonstrate that FeatureSORT also generalizes well beyond the MOT benchmarks. As shown in Table~\ref{tab:dancetrack}, FeatureSORTv7 surpasses the strong baseline ByteTrack by +2.6 MOTA, +8.4 HOTA, and +3.3 IDF1, highlighting its robustness under challenging conditions.

\begin{table}[h]
    \caption{Dance: Performance comparison with SOTA.}
    \label{tab:dancetrack}
    \centering
    \small
    \begin{tabular}{l c c c}
        \toprule
        Method & MOTA$\uparrow$ & HOTA$\uparrow$ & IDF1$\uparrow$ \\
        \hline
        TraDes \cite{wu2021track} & 86.2 & 43.3 & 41.2 \\
        SGT \cite{hyun2205detection} & 85.1 & 43.7 & 47.2 \\
        DeepSORT \cite{wojke2017simple} & 88.2 & 46.4 & 51.7 \\
        FairMOT \cite{zhang2021fairmot} & 82.2 & 39.7 & 40.8 \\
        ByteTrack \cite{zhang2022bytetrack} & 89.6 & 47.7 & 53.9 \\
        StrongSORT \cite{du2023strongsort} & 91.1 & 55.6 & \textbf{55.2} \\
        \hline
        \rowcolor{gray!20}
        FeatureSORTv7 & \textbf{92.2} & \textbf{56.1} & 55 \\
        \rowcolor{gray!20}
        FeatureSORTv8 & 91 & 55.4 & 54.5 \\
        \rowcolor{gray!20}
        FeatureSORTv7 & 91.6 & 55.3 & \textbf{55.2} \\
        \hline
    \end{tabular}
\end{table}

\section{Discussion}
\label{sec:discussion}
A known drawback of DeepSORT-like trackers is their computational cost, as they rely on a separate ReID network in addition to the detector, which slows inference. 
In FeatureSORT, while we also employ a separate ReID network for stronger embeddings, the redesigned YOLOX detector allows tracking to proceed even without it, making the system closer to joint detection–tracking approaches. 
Nevertheless, features obtained directly from the detector are generally less specialized than those extracted by dedicated networks for color, style, or direction classification. 
This introduces a trade-off between efficiency and accuracy: reducing reliance on external ReID modules improves speed, but may limit the discriminative power of features. 
FeatureSORT therefore requires balancing computational complexity, runtime performance, and tracking accuracy depending on the deployment scenario.

\section{Conclusion}
\label{sec:conclusion}
In this work, we introduced FeatureSORT, a modular and feature-enriched tracker designed to advance online multi-object tracking. 
The core contribution lies in redesigning the YOLOX detector to jointly predict bounding boxes and auxiliary attributes such as clothing color, clothing style, and movement direction. 
These outputs are incorporated into plug-and-play feature modules, enabling more robust and discriminative association. 
Through this redesign, FeatureSORT establishes itself as a strong and versatile baseline that unifies the strengths of joint detection–tracking and TBD paradigms. 
Extensive experiments on MOT16, MOT17, MOT20 and DanceTrack benchmarks demonstrate that FeatureSORT achieves SOTA accuracy, highlighting the effectiveness of feature-enriched detection and modular design in advancing the MOT field.

 \bibliographystyle{IEEEtran}  
\bibliography{Reference.bib}

\newpage

 




\end{document}